\documentclass{article}
\usepackage{ijcai25}
\usepackage[dvipsnames]{xcolor}
\usepackage{times}
\usepackage{soul}
\usepackage{url}
\usepackage[hidelinks]{hyperref}
\usepackage[utf8]{inputenc}
\usepackage[small]{caption}
\usepackage{graphicx}
\usepackage{amsmath}
\usepackage{amsthm}
\usepackage{amssymb}
\usepackage{booktabs}
\usepackage{algorithm}
\usepackage{algorithmic}
\usepackage[switch]{lineno}
\usepackage{float}
\usepackage{graphicx}
\usepackage{tikz}
\usepackage{forest}
\usepackage{amssymb} 
\usetikzlibrary{trees,positioning,shapes,shadows,arrows.meta}

\title{Empowering LLMs with Logical Reasoning: A Comprehensive Survey}

\author{
Fengxiang Cheng$^1$
\and
Haoxuan Li$^{2,3,4}$\thanks{Haoxuan Li and Zhouchen Lin are the corresponding authors.}\and
Fenrong Liu$^{3,1}$\and
Robert van Rooij$^1$\and\\
Kun Zhang$^{4,5}$\And
Zhouchen Lin$^{6,7,8*}$
\\
\affiliations
$^1$Institute for Logic, Language and Computation, University of Amsterdam\\
$^2$Center for Data Science, Peking University\\
$^3$Tsinghua-UvA JRC for Logic, Department of Philosophy, Tsinghua University\\ 
$^4$Machine Learning Department, MBZUAI\\
$^5$Department of Philosophy, CMU\\
$^6$State Key Lab of General AI, School of Intelligence Science and Technology, Peking University\\
$^7$Institute for Artificial Intelligence, Peking University\\
$^8$Pazhou Laboratory (Huangpu), Guangzhou
\emails
\{f.cheng, r.a.m.vanrooij\}@uva.nl,
hxli@stu.pku.edu.cn,
fenrong@tsinghua.edu.cn,\\
kunz1@cmu.edu,
zlin@pku.edu.cn
}
\begin{document}

\maketitle

\begin{abstract}

Large language models (LLMs) have achieved remarkable successes on various tasks. However, recent studies have found that there are still significant challenges to the logical reasoning abilities of LLMs, which can be categorized into the following two aspects: (1) \textbf{Logical question answering}: LLMs often fail to generate the correct answer \emph{within} a complex logical problem which requires sophisticated deductive, inductive or abductive reasoning given a collection of premises. (2) \textbf{Logical consistency}: LLMs are prone to producing responses contradicting themselves \emph{across} different questions. For example, a state-of-the-art question-answering LLM Macaw, answers \emph{Yes} to both questions \emph{Is a magpie a bird?} and \emph{Does a bird have wings?} but answers \emph{No} to \emph{Does a magpie have wings?}.
To facilitate this research direction, we comprehensively investigate the most cutting-edge methods and propose a detailed taxonomy. Specifically, to accurately answer complex logic questions, previous methods can be categorized based on reliance on external solvers, prompts, and fine-tuning. To avoid logical contradictions, we discuss concepts and solutions of various logical consistencies, including implication, negation, transitivity, factuality consistencies, and their composites. In addition, we review commonly used benchmark datasets and evaluation metrics, and discuss promising research directions, such as extending to modal logic to account for uncertainty and developing efficient algorithms that simultaneously satisfy multiple logical consistencies.












\end{abstract}

\section{Introduction}



Large language models (LLMs) have demonstrated remarkable performance in a broad range of natural language tasks including language generation, classification and translation. However, recent studies have found that there are still significant challenges to the logical reasoning abilities of LLMs. On the one hand, learning syntax, semantics, and world knowledge through tasks such as next-word prediction or masked language modeling does not ensure the logical reasoning ability of LLMs~\cite{luo2023towards}. 
On the other hand, the pre-training corpus of LLMs primarily consists of human-written texts, which lack high-quality logical reasoning samples such as logical deduction and proofs~\cite{morishitaenhancing}. These challenges significantly limit the applicability of LLMs due to the following two summarized aspects.  


\begin{figure*}[t!]
\centering
\resizebox{\textwidth}{!}{
\tikzset{
    basic/.style  = {draw, text width=2cm, align=center, rectangle},
    root/.style   = {basic, rounded corners=2pt, thin, align=center, fill=purple!10, text width=0.5cm,},
    node1/.style = {basic, thin, rounded corners=2pt, align=center, fill=green!07, text width=2cm,},
    node11/.style = {basic, thin, rounded corners=2pt, align=center, fill=green!04, text width=2cm,},
    node12/.style = {basic, thin, rounded corners=2pt, align=center, fill=green!04, text width=2cm,},
    node13/.style = {basic, thin, rounded corners=2pt, align=center, fill=green!04, text width=2cm,},
    node14/.style = {basic, thin, rounded corners=2pt, align=center, fill=green!04, text width=2cm,},
    node111/.style = {basic, thin, rounded corners=2pt, align=center, fill=green!01, text width=13cm,},
    node121/.style = {basic, thin, rounded corners=2pt, align=center, fill=green!01, text width=13cm,},
    node131/.style = {basic, thin, rounded corners=2pt, align=center, fill=green!01, text width=13cm,},
    node141/.style = {basic, thin, rounded corners=2pt, align=center, fill=green!01, text width=13cm,},
    node2/.style = {basic, thin, rounded corners=2pt, align=center, fill=orange!20, text width=2cm,},
    node21/.style = {basic, thin, rounded corners=2pt, align=center, fill=orange!15, text width=2cm,},
    node22/.style = {basic, thin, rounded corners=2pt, align=center, fill=orange!15, text width=2cm,},
    node23/.style = {basic, thin, rounded corners=2pt, align=center, fill=orange!15, text width=2cm,},
    node24/.style = {basic, thin, rounded corners=2pt, align=center, fill=orange!15, text width=2cm,},
    node25/.style = {basic, thin, rounded corners=2pt, align=center, fill=orange!15, text width=2cm,},
    node211/.style = {basic, thin, rounded corners=2pt, align=center, fill=orange!10, text width=2cm,},
    node212/.style = {basic, thin, rounded corners=2pt, align=center, fill=orange!10, text width=2cm,},
    node2111/.style = {basic, thin, rounded corners=2pt, align=center, fill=orange!5, text width=10.05cm,},
    node2121/.style = {basic, thin, rounded corners=2pt, align=center, fill=orange!5, text width=10.05cm,},
    node221/.style = {basic, thin, rounded corners=2pt, align=center, fill=orange!5, text width=13cm,},
    node231/.style = {basic, thin, rounded corners=2pt, align=center, fill=orange!5, text width=13cm,},
    node241/.style = {basic, thin, rounded corners=2pt, align=center, fill=orange!5, text width=13cm,},
    node251/.style = {basic, thin, rounded corners=2pt, align=center, fill=orange!5, text width=13cm,},
    edge from parent/.style={draw=black, edge from parent fork right}
}
\begin{forest} for tree={
    grow=east,
    growth parent anchor=west,
    parent anchor=east,
    child anchor=west,
    edge path={\noexpand\path[\forestoption{edge}]  (!u.parent anchor) -- +(10pt,0pt) |-  (.child anchor) \forestoption{edge label};}
    l sep=7mm,
    calign=center,
}
[\rotatebox{90}{Logical Reasoning}, root, l sep=7mm, yshift = -11mm
    [Logical Consistency, node2, l sep=7mm, 
        [Compositional, node25,l sep=5mm, yshift = 0.1mm
            [{REFLEX~\shortcite{kassner-etal-2023-language}, REPAIR~\shortcite{liu2024aligning}, LoCo-LMs~\shortcite{calanzone2024logically}, LLMQuery~\shortcite{ghosh2025logical}},node251, edge path={\noexpand\path[\forestoption{edge}] (!u.parent anchor) --  (.child anchor) \forestoption{edge label};}]
        ]
        [Factuality, node24,l sep=5mm, yshift = 0.2mm
            [{BeliefBank~\shortcite{kassner-etal-2021-beliefbank}, LoCo-LMs~\shortcite{calanzone2024logically}, LLMQuery~\shortcite{ghosh2025logical}},node241, edge path={\noexpand\path[\forestoption{edge}] (!u.parent anchor) --  (.child anchor) \forestoption{edge label};}]
        ]
        [Transitivity, node23,l sep=5mm, yshift = 0.1mm
            [{Logic-guided Consistency Regularization~\shortcite{asai-hajishirzi-2020-logic}, ConCoRD~\shortcite{mitchell-etal-2022-enhancing}, REPAIR~\shortcite{liu2024aligning}},node231, edge path={\noexpand\path[\forestoption{edge}] (!u.parent anchor) --  (.child anchor) \forestoption{edge label};}]
        ]
        [Implication, node22,l sep=5mm, yshift = -3.8mm
            [{BeliefBank~\shortcite{kassner-etal-2021-beliefbank}, ConCoRD~\shortcite{mitchell-etal-2022-enhancing}, Maieutic Prompting~\shortcite{jung-etal-2022-maieutic}, 
            \\REFLEX~\shortcite{kassner-etal-2023-language}, REPAIR~\shortcite{liu2024aligning}, 
            \\LoCo-LMs~\shortcite{calanzone2024logically},  LLMQuery~\shortcite{ghosh2025logical} },node221, edge path={\noexpand\path[\forestoption{edge}] (!u.parent anchor) --  (.child anchor) \forestoption{edge label};}]
        ]
        [Negation, node21,l sep=7mm
            [Antonyms, node212,l sep=5mm, yshift = 0.2mm
                [{Logic-guided Consistency Regularization~\shortcite{asai-hajishirzi-2020-logic}, REPAIR~\shortcite{liu2024aligning}}, node2121, edge path={\noexpand\path[\forestoption{edge}] (!u.parent anchor) --  (.child anchor) \forestoption{edge label};}]
            ]
            [Negation, node211,l sep=5mm,  yshift = -3.7mm
                [{BeliefBank~\shortcite{kassner-etal-2021-beliefbank}, ConCoRD~\shortcite{mitchell-etal-2022-enhancing}, \\Maieutic Prompting~\shortcite{jung-etal-2022-maieutic}, REFLEX~\shortcite{kassner-etal-2023-language}, REPAIR~\shortcite{liu2024aligning}, \\LoCo-LMs~\shortcite{calanzone2024logically}, LLMQuery~\shortcite{ghosh2025logical}},node2111, edge path={\noexpand\path[\forestoption{edge}] (!u.parent anchor) --  (.child anchor) \forestoption{edge label};}]
            ]
        ]
    ]
    [Logical Question Answering, node1,   l sep=7mm
        [Benchmark, node11, l sep=5mm,  yshift = -4.1mm
            [{LogicNLI~\shortcite{tian2021diagnosing}, LogiGLUE~\shortcite{luo2023towards}, LogicBench~\shortcite{parmar-etal-2024-logicbench}, 
LINK~\shortcite{li2024search}, LogicAsker~\shortcite{wan-etal-2024-logicasker}, 
Multi-LogiEval~\shortcite{patel2024multi},
LogiEval~\shortcite{liu2025evaluating},
LogicPro~\shortcite{jiang2024logicpro}, 
AutoLogi~\shortcite{zhu2025autologi},
SATBench~\shortcite{wei2025satbench},
SmartyPat-Bench~\shortcite{xu2025socrates} }, 
            node111, edge path={\noexpand\path[\forestoption{edge}] (!u.parent anchor) --  (.child anchor) \forestoption{edge label};}]
        ]
        [Fine-tuning, node12,   l sep=5mm, yshift = -1.8mm
            [{LReasoner~\shortcite{wang-etal-2022-logic}, DiLA~\shortcite{zhang2024dilaenhancingllmtool}, AMR-LDA~\shortcite{bao-etal-2024-abstract},  LoGiPT~\shortcite{feng-2024-languagecanbe}, ALT~\shortcite{morishitaenhancing}, LogicAsker~\shortcite{wan-etal-2024-logicasker}, LogicLLM~\shortcite{jiao2024exploring}, Unigram~\shortcite{sileo-2024-scaling}}, node121, edge path={\noexpand\path[\forestoption{edge}] (!u.parent anchor) --  (.child anchor) \forestoption{edge label};}]
        ]        
        [Prompt-based, node13,   l sep=5mm, yshift = -3.7mm
            [{CR~\shortcite{zhang2023cumulative}, DoT~\shortcite{zhang2024diagram}, \\SymbCoT~\shortcite{xu-etal-2024-faithful}, LINA~\shortcite{li2024leveraging},  ChatLogic~\shortcite{wang2023chatlogic}, NeuBAROCO~\shortcite{ozeki-etal-2024-exploring}, Aristotle~\shortcite{xu2025aristotle}, LoT~\shortcite{liu2024logicofthoughtinjectinglogiccontexts}}, node131, edge path={\noexpand\path[\forestoption{edge}] (!u.parent anchor) --  (.child anchor) \forestoption{edge label};}]
        ]
        [Solver-based, node14,  l sep=5mm,  yshift = -2mm
            [{SatLM~\shortcite{ye2023satlm}, Logic-LM~\shortcite{pan-etal-2023-logic}, LINC~\shortcite{olausson-etal-2023-linc}, $\forall$uto$\exists$val~\shortcite{karia2024forall}, CLOVER~\shortcite{ryu2025divide}, VERUS-LM~\shortcite{callewaert2025verus}}, node141, edge path={\noexpand\path[\forestoption{edge}] (!u.parent anchor) --  (.child anchor) \forestoption{edge label};}]
        ]
    ]
]
\end{forest}
}
\caption{Our taxonomy tree of Logical Reasoning, which is primarily divided into Logical Question Answering and Logical Consistency.}
\label{fig:lit_surv}
\end{figure*}

LLMs often fail to generate the correct answer in \textbf{logical question answering}, which requires sophisticated deductive, inductive or abductive reasoning given a collection of premises and constraints. Specifically, these logical questions can be broadly divided into two categories: (1) Determine whether a statement can be deduced from the given information, namely, output the truth value of the statement: true, false or unknown. For example, the premise and constraints could be \emph{Metals conduct electricity. 
If something is made of iron, then it is metal. Nails are made of iron.}, followed by the logical question \emph{Is the following statement true, false, or unknown? Nails cannot conduct electricity}. To answer this question correctly, LLMs need to conduct logical reasoning \emph{nails$\to$made of iron$\to$metal$\to$conduct electricity} before concluding that the statement is actually \emph{false}. (2) Find the correct option that can satisfy all the given premises from the multiple choices. Surprisingly, LLaMA-13B achieves 33.63\% accuracy under 8-shot prompting on the logical questions dataset FOLIO, which is only slightly better than a random guess from true, false and unknown with an accuracy of 33.33\%~\cite{han-etal-2024-folio}. This significantly restricts the application of LLMs in complicated real-world situations, such as problem-solving and decision-making.

LLMs are also prone to producing responses contradicting themselves \emph{across} different questions, which is regarded as a violation of \textbf{logical consistency}. Note that the form of logical consistency can be diverse. For example, LLaMA-2-70B answers \emph{true} to both questions \emph{Is an albatross an organism?} and \emph{Is an albatross not an organism?}~\cite{calanzone2024logically}, which violates the negation consistency (see Section \ref{sec:3} for formal definition). In addition, a state-of-the-art Macaw question-answering LLM answers \emph{Yes} to both questions \emph{Is a magpie a bird?} and \emph{Does a bird have wings?} but answers \emph{No} to \emph{Does a magpie have wings?}~\cite{mitchell-etal-2022-enhancing}, which violates the transitivity consistency. Unfortunately, many studies have shown that training on large question-answering datasets alone cannot ensure the logical consistency of LLMs~\cite{kassner-etal-2021-beliefbank,jung-etal-2022-maieutic}. As a result, these contradictory outputs concern the reliability and trustworthiness of LLMs, which limits the practical deployment in especially high-stakes scenarios.




In this paper, we consider accurately answering isolated complex logical questions and ensuring logical consistency across outputs to different questions as two sides of the coin 
in improving the logical reasoning capabilities of LLMs. 
We comprehensively investigate the most cutting-edge methods for both and propose a detailed taxonomy, as shown in Figure \ref{fig:lit_surv}. Specifically, for logical question answering, these methods are classified into solver-based, prompt-based, and fine-tuning methods. In general, solver-based methods translate natural language problems to symbolic language expressions, and then solve them via external logical solvers~\cite{olausson-etal-2023-linc}. Prompt-based methods either explicitly model the logical chain when answering the questions~\cite{zhang2024diagram}, or translate natural language into symbolic language by well-defined prompts, and then utilize LLMs for better reasoning~\cite{xu-etal-2024-symbol}. 
Fine-tuning methods train LLMs with augmented deductive proofs and natural language examples explicitly including the logical reasoning process. For logical consistency, we formulate the most common types of violations, including negation, implication, transitivity, factuality consistencies, and their composites, discuss the state-of-the-art solutions, and review commonly used benchmark datasets and evaluation metrics. Lastly, we discuss promising research directions, such as extending to modal logic to account for uncertainty and developing efficient algorithms satisfying multiple logical consistencies.


To the best of our knowledge, this is the first work to comprehensively investigate the most cutting-edge research on improving LLM logical reasoning capabilities, covering complex logical question answering and logical consistency.  A relevant survey work~\cite{zong-lin-2024-categorical} focuses on categorical syllogisms, as a specific logical reasoning paradigm, but ignores other complex logic inference rules. In addition, another survey~\cite{lam2024closerlooklogicalreasoning} only discuss the methods based on external tools such as logical solvers.

\begin{figure*}[t!]
    \centering
    \includegraphics[width=\linewidth]{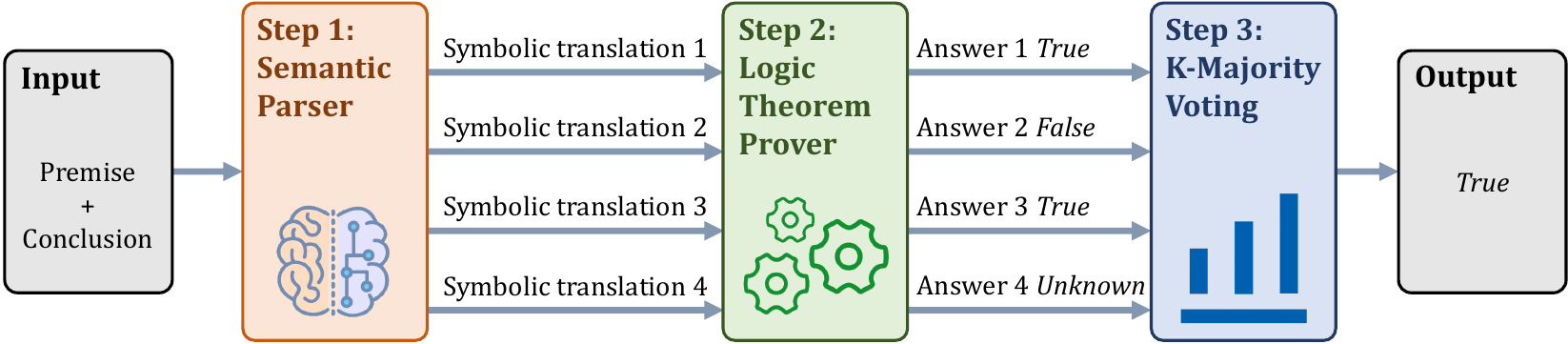}
    \caption{The overview of workflow in solver-based methods for logical question answering.}
    \label{fig:solver_framework}
\end{figure*}

\begin{figure}
    \centering
    \includegraphics[width=0.7\linewidth]{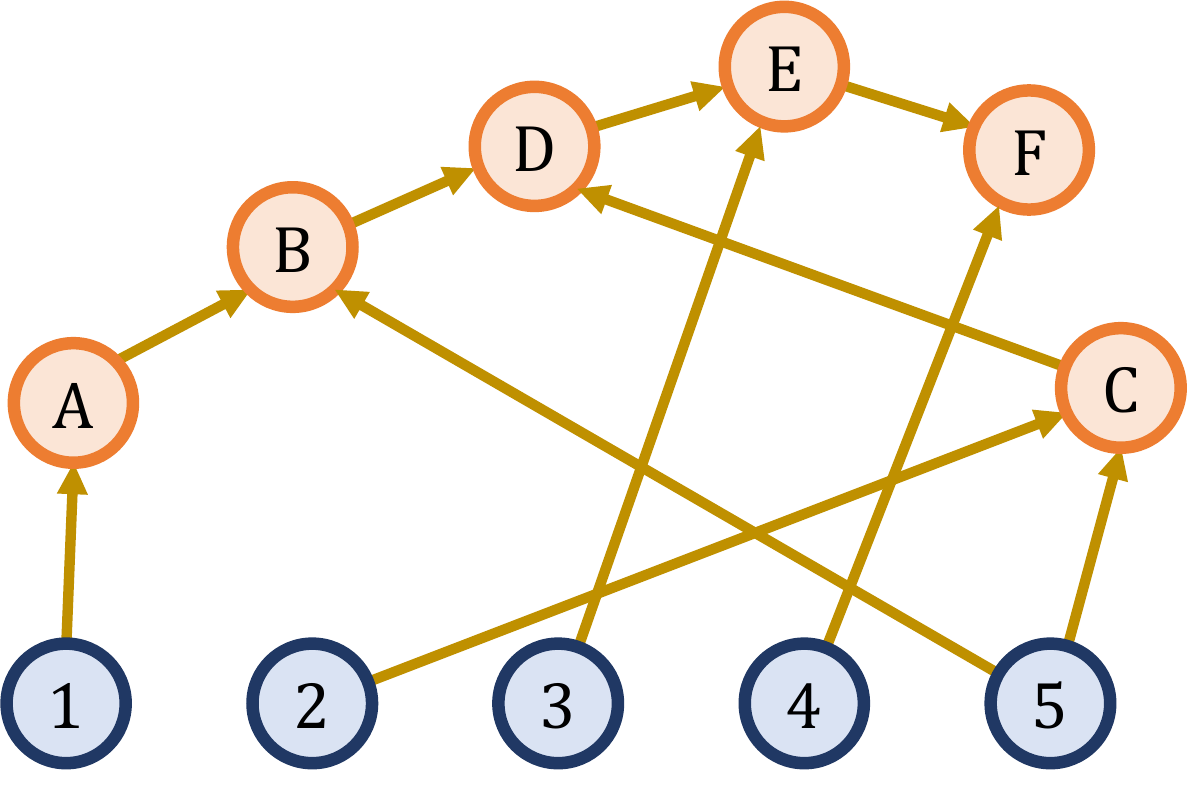}
    \caption{The logical derivation process could be represented as a DAG, where 1-5 represent premises, A-F represent the  derivation proceed and every edge represents an individual inference.}
    \label{fig:enter-label}
    \vspace{-6pt}
\end{figure}

\section{Logical Question Answering}

Leveraging LLMs for complex reasoning problems has been a key focus of recent research~\cite{luo2023towards,sun-etal-2024-determlr,zhang2024diagram}. Numerous studies have explored methods to enhance the logical reasoning capabilities of LLMs, which can be broadly categorized into solver-based (Sec. \ref{sec:2.1}), prompt-based (Sec. \ref{sec:2.2}), and fine-tuning methods (Sec. \ref{sec:2.3}). In addition, many benchmarks are developed to evaluate the LLM logical reasoning capabilities (Sec. \ref{sec:2.4}).

\subsection{Solver-Based Methods}\label{sec:2.1}

The solver-based methods translate \textbf{natural language (NL)} questions to \textbf{symbolic language (SL)} expressions, and then solve them via external solvers. The workflow is shown in Figure \ref{fig:solver_framework}, including the following three steps. First, these methods use LLMs to translate NL into SL (such as logic programming (LP), first-order logic (FOL), constraint satisfaction (CSP), or boolean satisfiability (SAT) formulation) that can be recognized by corresponding solvers. Next, they adopt an external solver for logical reasoning to output the desired answer. Finally, LLMs are used to translate the SL answer provided by the solver into NL to generate the final answer based on ensemble algorithms such as majority vote.

Specifically, Faithful Chain-of-Thought~\cite{lyu2023faithfulCoT} translates an NL query into an SL reasoning chain with a deterministic solver (e.g., Python/Datalog interpreter) to improve the faithfulness of LLM-generated answers. 
Focusing on logical reasoning motivated by this approach, SatLM~\cite{ye2023satlm} and LINC~\cite{olausson-etal-2023-linc} propose to integrate logical solvers into the reasoning processes of LLMs. In particular, LINC leverages prompting to enable LLMs to translate NL inputs comprising multiple premises and a conclusion into SL expressions. To mitigate translation errors that arise from LLMs, LINC generates multiple SL samples for a single NL problem. The solver then produces the corresponding results as candidate outputs, after which k-majority voting is applied to select the optimal output as the final answer to the logic reasoning question. SatLM exploits the LLM to translate the NL question into a set of logical constraints and then employs an SAT solver to execute the proof plan and derive answers. Unlike LINC and SatLM adopt a single type of SL and use its corresponding solver, LogicLM~\cite{pan-etal-2023-logic} utilizes LLMs to translate NL problems into a task-specific formulation, which could be the LP language, FOL, CSP, and SAT formulation, tailored to different datasets. LogicLM then utilizes the corresponding solvers to obtain the answers and finally translates them back to NL. 

While the above methods demonstrate remarkable performance, they heavily rely on the correctness of the translation between NL and SL. To enhance translation accuracy, CLOVER~\cite{ryu2025divide} first translates the raw NL paragraph to atomic NL subsentences with their logical dependency structure, then translates to the target SL. VERUS-LM~\cite{callewaert2025verus} introduces a self-refinement step that uses feedback from the reasoning engine to correct erroneous logical statements, performing both syntactic refinement and semantic refinement. $\forall$uto$\exists$val~\cite{karia2024forall} adds a bidirectional translation loop of SL$\rightarrow$NL (interpretation) and NL$\rightarrow$SL (compilation), and then uses logical solvers to automatically check the logical equivalence of SL expressions before and after the translation loop, to ensure the correctness of LLM answers without human annotation.

The limitations of the solver-based methods can be summarized as follows. First, transforming logical questions into formal expressions will result in information loss~\cite{li2024leveraging,liu2024logicofthoughtinjectinglogiccontexts}, leading to unsolvable problems.
For example, in the symbolic translation of ``When a person reads a book, that person gains knowledge. \textit{Harry} reads \textit{Walden}. Whether this inference is correct: Harry gains knowledge", the solver will output uncertain, since the symbolic translation process loses the vital hidden information ``Harry is a person" and ``Walden is a book", which can be easily inferred by humans~\cite{liu2024logicofthoughtinjectinglogiccontexts}. 
Second, a small mistake in translation from the NL questions to the SL formulation will severely affect the results derived from SL solvers. For example, GPT 3.5 is not able to add a complete bracket in a logical formula, leading to wrong SL with different meanings~\cite{feng-2024-languagecanbe,lam2024closerlooklogicalreasoning}. Lastly, as the complexity of the question increases, the search space expands exponentially, decreasing both the effectiveness and efficiency~\cite{zhang2024dilaenhancingllmtool}.

\subsection{Prompt-Based Methods}\label{sec:2.2}

Prompting is a direct and effective technique for eliciting the logical reasoning capabilities of LLMs, which can be broadly divided into two categories. The first is to explicitly model the logical chain when answering the questions. For example, the Chain-of-Thought (CoT) \cite{wei2022chain} prompting strategy enables LLMs to output the reasoning process step-by-step. Based on this, Tree-of-Thought (ToT)~\cite{yao2024tree} is proposed to let LLMs self-evaluate and select among multiple inference paths. Graph-of-Thought (GoT)~\cite{besta2024GoT} enhances LLMs’ capabilities through networked reasoning, by modeling LLM thought as a vertex and representing the dependencies between such thoughts via edges.
To better mirror human reasoning process, as shown in Figure~\ref{fig:enter-label}, cumulative reasoning (CR)~\cite{zhang2023cumulative}  decomposes the complex problem into small and manageable components, and modeling the logical derivation process as a direct acyclic graph (DAG) among these components, in which previous propositions are leveraged for effective composition. Diagram-of-Thought (DoT)~\cite{zhang2024diagram} prompts a single LLM to construct and navigate the DAG by role-specific tokens (e.g., proposer, critic, summarizer).

The second category is to obtain the symbolic expressions and to leverage its advantages by well-defined LLMs' prompts. For example, SymbCoT~\cite{xu-etal-2024-symbol} prompts LLMs to first translate NL problems to SL formulation, then generate a step-by-step solution to address the task with logic inference rules such as modus ponens (MP) rule, and finally to verify the correctness of the translation and answers. Instead of using linguistic token relations for decomposing logical problems, which may lead to disconnected sub-problems and faulty reasoning, the Aristotle framework~\cite{xu2025aristotle} proposes to exploit the underlying logical structure for decomposition to improve both efficacy and efficiency. In addition, Logic-of-Thought (LoT)~\cite{liu2024logicofthoughtinjectinglogiccontexts} instructs LLMs to translate and expands the implicated logical expressions based on the logic rules, then adding these expanded logical information into the input prompt to derive the final answers. LINA~\cite{li2024leveraging} design LLMs' prompts via hypothetical-deductive reasoning paradigm to reduce the expansive search space compared to the traditional forward reasoning methods. ChatLogic~\cite{wang2023chatlogic} integrates symbolic logic programming (e.g., pyDatalog) through interactive prompts, which corrects semantic and syntactic errors of the programming to improve the deductive accuracy. 

Despite the transparency and interpretability of prompt-based methods, they still have the following limitations. First, despite the prompt-based methods can correctly translate from NL to SL, reasoning errors remain, suggesting that
the source of these errors is not the misinterpretation of the premises but rather the reasoning process itself~\cite{ozeki-etal-2024-exploring}. Second, compared with solver-based methods, the prompt-based methods do rely more on the LLM's own reasoning capabilities, leading to sub-optimal performance when facing the complex reasoning tasks. Lastly, solving a problem typically requires numerous inference steps, in which the LLMs are iteratively prompted with various search strategies~\cite{yao2024tree}, resulting in a significant computational cost~\cite{yang2023neuro}.




\begin{figure*}
    \centering
    \includegraphics[width=\linewidth]{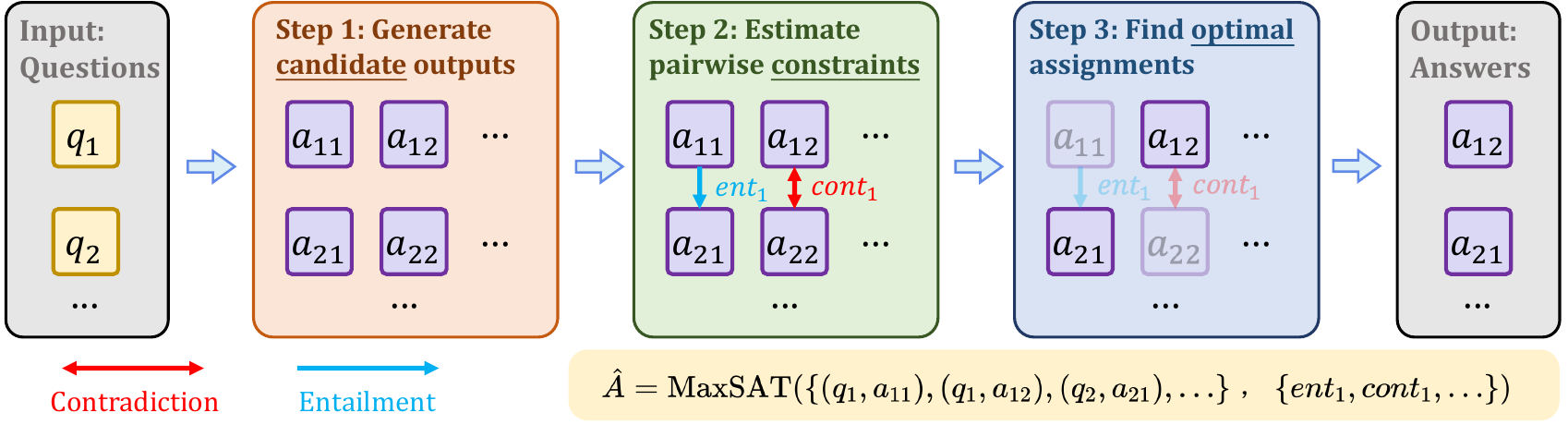}
    \caption{A general framework to enhance the logical consistency of LLM outputs across different questions.}
    \label{fig:figure4}
    \vspace{-6pt}
\end{figure*}

\subsection{Fine-Tuning Methods}\label{sec:2.3}
The limited reasoning abilities of LLMs can be attributed to the lack of high-quality reasoning samples (especially logical multi-step deduction or proofs) in the pre-training corpus, which is composed mainly of human-written texts~\cite{morishitaenhancing}. Human-written texts usually show reflexive thinking rather than rigid reasoning, let alone logical deduction or inference. Intuitively, training LLMs with deductive proofs and NL examples including the explicit logical reasoning process provides an effective way to enhance the intrinsic logical reasoning capabilities of LLMs.

In general, the fine-tuning data for enhancing the logical reasoning capabilities of LLMs are generated via logic rules. Specifically, 
LogicAsker~\cite{wan-etal-2024-logicasker} formally defines a comprehensive set of 34 atomic logic rules and 208 extended rules necessary for LLMs to execute formal reasoning based on propositional and predicate logic. By evaluating reasoning ability on each rule, LogicAsker figures out the reasoning rules that LLM performs poorly, based on which creating corresponding in-context-learning examples and fine-tuning data to improve reasoning abilities. Similar to LogicAsker, ALT~\cite{morishitaenhancing} builds a synthetic logic corpus based on diverse logical reasoning rules such as syllogism, contraposition, and De Morgan’s laws, comprising numerous samples of multi-step deduction with unknown facts, diverse linguistic expressions, and challenging distractors. Instead of directly augmenting NL via logical rules, AMR-LDA~\cite{bao-etal-2024-abstract} first converts the original NL into an Abstract Meaning Representation (AMR) graph to capture the logical structure of the sentence, then augments NL via logically augmented AMR graphs. In addition, incorporating deeper reasoning steps for fine-tuning the LLMs can further enhance the logical reasoning capabilities of LLMs~\cite{bao2022multi}.

The effectiveness of fine-tuning methods can be further improved by incorporating more symbolic reasoning process and challenging instances, as well as solution refinement. Specifically, LoGiPT~\cite{feng-2024-languagecanbe} 
empowers LLMs by directly
learning the reasoning process of logical solvers, avoiding the risk of no answer to the logical questions when facing parsing errors for the solver-based methods. LReasoner~\cite{wang-etal-2022-logic} proposes to augment challenging instances with literally similar but logically different instances and incorporates contrastive learning to better capture logical information such as logical negative and conditional relationships. DiLA~\cite{zhang2024dilaenhancingllmtool}
utilizes a LLM to generate an original solution, and then iteratively refines it by forward and backward passes through a network logic layer involving first-order logic constraints. Moreover, Unigram~\cite{sileo-2024-scaling} shows that simple declarative grammars paired with solvers can outperform complex proof tree generators for reasoning dataset generations. To reduce manual annotation costs, LogicLLM~\cite{jiao2024exploring}
introduces a fully self-supervised framework for integrating logical reasoning capabilities into LLMs.

\subsection{Evaluation: Tasks and Benchmark Datasets}\label{sec:2.4}

Typically, a logical reasoning question contains an NL description that outlines a set of propositions or constraints related to certain objects, along with a corresponding question about these objects. The objective is to infer the correct answer to the question based on the given information~\cite{ye2023satlm}. These question-answering pairs can be typically categorized into two primary types~\cite{luo2023towards}:

(1) Logical Statement Checking: Assess whether a given question or statement can be logically inferred from the provided information, producing an output choosing from true, false, or unknown. The formulation includes logic programming, propositional logic, and first-order logic language. Typical datasets include Proofwriter~\cite{tafjord2021proofwriter}, FOLIO~\cite{han-etal-2024-folio}, Ruletaker~\cite{clark2021transformers}, etc.


(2) Multiple Choice Question-Answering: Identify the correct option from multiple choices satisfying all given premises and constraints. The formulation includes constraint satisfaction problems and Boolean SAT formulation.  Typical datasets include ReClor~\cite{WeihaoReClor2020}, LogiQA~\cite{liu2021logiqa}, AR-LSAT~\cite{zhong2022analytical}, etc.


In addition, several studies have established benchmarks for evaluating LLMs' various types logical reasoning. Specifically, LogicNLI~\cite{tian2021diagnosing} assesses FOL reasoning via natural language inference. LogiGLUE~\cite{luo2023towards} and LogiEval~\cite{liu2025evaluating} cover deductive, abductive, inductive, and analogical reasoning. LogicPro~\cite{jiang2024logicpro} synthesizes complex logical data using algorithm problems with Python code, while LINK~\cite{li2024search} systematically generates long-tail reasoning data via logical rules. LogicBench~\cite{parmar-etal-2024-logicbench} and Multi-LogiEval~\cite{patel2024multi} consist of propositional, first-order, and non-monotonic logic, in which Multi-LogiEval contains samples with reasoning steps up to five. AutoLogi~\cite{zhu2025autologi} and SATBench~\cite{wei2025satbench} use logical puzzles to create datasets with varying difficulties, while SmartyPat-Bench~\cite{xu2025socrates} generates annotated logical fallacies using Prolog rules.

\section{Logical Consistency}\label{sec:3}

Logical consistency requires that LLM responses to different questions do not conflict with each other or with established knowledge bases, in accordance with logical principles~\cite{mitchell-etal-2022-enhancing,tafjord-etal-2022-entailer}. This prevents LLMs from contradicting themselves in multiple outputs and is consistent with external knowledge~\cite{daniel2025check,ghosh2025logical}, improving the reliability and trustworthiness especially in high-stakes real-world deployment so as to strengthening end-users' experience and confidence.

Unfortunately, merely training on QA datasets is insufficient to meet the consistency requirements. To address this issue, many methods have been proposed to improve the logical consistency of LLMs. Hereafter, we will take a closer look at various logical consistencies, including logical relationships among one, two, three statements and their composites. Meanwhile, we will summarize corresponding methods for enhancing LLM logical consistencies as well as the evaluation metrics.

\subsection{Negation Consistency} 

Negation consistency, as the basis of logical consistency, requires that $p$ and $\neg p$ cannot hold simultaneously while one of them is true $p\oplus \neg p$, \textit{i.e.}, $(p\vee\neg p)\wedge\neg(\neg p\wedge p)$~\cite{kassner-etal-2023-language}. In NL, there are two forms of negation consistency: the first is to directly add \emph{not} to the statement, and the second is to replace the original adjective in the statement with its antonym~\cite{asai-hajishirzi-2020-logic}, e.g., replacing \emph{heavier} with \emph{lighter}. Empirically, recent studies found that LLMs are prone to producing responses that contradict themselves across different questions~\cite{kassner-etal-2023-language,ghosh2025logical}. For example, LLaMA-2-70B answers true to both questions \textit{Is an albatross an organism?} and \textit{Is an albatross not an organism?}.

To address this challenge, Beliefbank~\cite{kassner-etal-2021-beliefbank} stores LLM's raw answers (or beliefs) and flips the beliefs that clash significantly
with others using a constraint solver (e.g., MaxSAT solver). The modified beliefs are then used as query context via a feedback mechanism for improving both consistency and accuracy of the LLM when facing relevant questions. Similarly, as shown in Figure \ref{fig:figure4}, ConCoRD~\cite{mitchell-etal-2022-enhancing} first generates several candidate outputs from
the initial model, then estimates soft pairwise inconsistencies between the outputs using natural language inference, and finally finds the most satisfactory output for each question via a MaxSAT solver.

\subsection{Implication Consistency} 
The implication consistency is based on the logical rule $p\to q, p \vDash q$. It means that given a constraint ``$p\to q$" and the premise $p$, it can be inferred that ``$q$ is True". If the model outputs that ``$q$ is False", then we say the output violates implication consistency. For example, given the physical fact that \emph{All Iron (p) is metal (q)}, LLMs are not to be expected to answer \emph{True} to \emph{This material is iron (p)} and \emph{False} to \emph{This material is metal (q)} at the same time.


Intuitively, implication consistency is closely related to the CoT method because each chain can be regarded as a logical implication. The incorrectness of LLM-generated explanations may lead to logically inconsistent and unreliable outputs. To infer a correct output to a question even from the unreliable generations of LLMs, Maieutic Prompting~\cite{jung-etal-2022-maieutic} offers a novel few-shot inference method that infers a correct answer by inducing the LLMs
to generate abductive explanations (e.g., X is true, because ...) for diverse hypotheses and enumerating a structure of
explanations — possibly noisy and contradictory — and resolving them with a symbolic inference algorithm. 

\subsection{Transitivity Consistency} 
Transitivity represents the logical relationships among three logical statements. Formally, given two premises $p\to q$ and  $q\to r $, it can be inferred that $p\to r$, which is regarded as the transitivity consistency. It has been shown that LLMs lack transitivity consistency. For example, the state-of-the-art QA model Macaw answers both \textit{Yes} to \textit{Is a sparrow a bird?} and \textit{Does a bird have feet?}, but answers \textit{No} to \textit{Does a sparrow have feet?}~\cite{mitchell-etal-2022-enhancing}. From the two affirmative answers, it can be implied that \textit{A sparrow has the feet} according to the transitivity rule, which is incompatible with the negative answer to \textit{Does a sparrow have feet?}.

To mitigate the transitivity inconsistency of LLMs, a logic-guided data augmentation and consistency regularization approach~\cite{asai-hajishirzi-2020-logic} leverages logical and linguistic knowledge to augment labeled training data and then uses a consistency-based regularizer to train the LLMs. For example, based on the questions \textit{If a tsunami happens, will wood be more moist?} and \textit{If wood is more moist, is more weathering occurring?} along with answers \textit{More} and \textit{More}, the implicated information \textit{If a tsunami happens, is more weathering occurring?} together with the answer \textit{More} will then be augmented by the transitivity. Similarly, for improving the negation consistency, the antonyms as well as the negation of the statements will also be added to the dataset. In this way, this approach achieves large improvements over previous methods in a variety of logical QA tasks.


\subsection{Factuality Consistency} 

Factuality consistency refers to the degree of alignment between the generated outputs from LLMs and the real-world knowledge base (KB). It is closely related to fact-checking tasks, which involve evaluating the factual accuracy of model outputs by comparing them with reliable knowledge sources, while also detecting potential logical errors or factual inaccuracies. Factuality consistency differs from other logical consistencies in that it requires LLMs not to violate an external knowledge base~\cite{calanzone2024logically}. In contrast, the latter requires that logical rules not be violated.


To address the issue of insufficient factuality consistency in existing LLMs under complex query scenarios, an evaluation and improvement framework is proposed through a combination of knowledge graph (KG) integration and supervised fine-tuning~\cite{ghosh2025logical}. Using a retrieval-augmented generation (RAG) framework, they provide LLMs with KG contexts (subgraphs extracted via breadth first search) to ground fact-checking queries. The key approach involves supervised fine-tuning on three datasets constructed in the RAG manner to enforce consistency, where the model learns to align responses with logical rules (e.g., commutative laws, negation consistency) and notably KG facts.
By leveraging structured KG facts and explicit logical constraints, the approach enhances LLMs’ factuality consistency in fact-checking tasks involving propositional logic. 
 

\subsection{Compositional Consistency} 


Compositional consistency refers to the ability to maintain various types of logical consistency when combining multiple facts or logical constraints. Specifically, when a model needs to combine independent facts into a complex chain of reasoning by logical operators (e.g., implication, conjunction, etc.), it should ensure that each step of the derivation conforms to the logical rules, and makes final conclusions self-consistent and logically correct to avoid contradictions. This requires LLMs not only understand the meaning of individual fact, but also to correctly capture the logical relationships when they are combined, avoiding LLMs' inference errors.

For LLMs' outputs to satisfy compositional consistency, LoCo-LMs~\cite{calanzone2024logically} introduces a loss based on neuro-symbolic reasoning that teaches an LLM to be logically consistent. The loss encourages the LLM to perform principled probabilistic reasoning at training time by
maximizing the probability of LLM's beliefs that comply with the provided set of logical constraints. In addition, REPAIR~\cite{liu2024aligning} proposes a universal framework to quantify the compositional logical consistency including three fundamental properties: \emph{transitivity}, \emph{commutativity}, and \emph{negation invariance}, then refines noisy pairwise comparisons using rank aggregation and augments logically consistent comparisons for instruction-tuning. This leads to improved logical consistency while maintaining alignment with human
preferences.



\subsection{Evaluation}
BeliefBank~\cite{kassner-etal-2021-beliefbank} introduces a gold-standard for evaluating the accuracy and consistency of LLM-generated answers, which has been widely adopted by subsequent work~\cite{mitchell-etal-2022-enhancing,kassner-etal-2023-language}. The evaluation framework typically includes:
\begin{itemize}
    \item A set of entities $S=\{s_m\}_{m=1}^M$;
    \item A set of unary predicates $P=\{P_n\}_{n=1}^N$;
    \item A collection of facts $\{\left(P_n\left(s_m\right)\right)_i\}_{i=1}^{I}$, whose binary truth value $\{0,1\}$ is known;
    \item A directed graph of constraints $G(P, E)$, whose edges $\left(P_n, P_{n^{\prime}}\right) \in$ $E$ represent  $\forall s_m\left(P_n(s_m) \rightarrow P_{n^{\prime}}(s_m)\right)$.
\end{itemize}

From these, the simple yes/no questions are constructed using NL templates. For example, for fact $P_n\left(s_m\right)$, if entity $s_m$ represents a student and predicate $P_n$ represents an ability to play the piano, then the corresponding template-generate QA pair ($q_i, a_i$) means the question \textit{$Q$: Is it true that a student is able to play piano?} together with the answer $A$: \textit{Yes} or \textit{No}.


To assess the logical consistency within LLM-generated answers across different questions, Beliefbank~\cite{kassner-etal-2021-beliefbank}, ConCoRD~\cite{mitchell-etal-2022-enhancing} and many subsequent work apply the complement of \emph{conditional constraint violation metric} $\tau$~\cite{li2019logic}, that is, $1-\tau$, in which $\tau$ is defined as the proportion of \emph{relevant} constraints in $G$ which are \emph{violated}. Specifically, we say a constraint $\forall x\left(P_n(x) \rightarrow P_{n^{\prime}}(x)\right)$ is relevant if for some entity $s_m$, there exists some LLM outputs considering the premise $\left( P_n(s_m) \right)_i$ is true, and there also exists some LLM outputs corresponding to the conclusion $\left( P_{n'}(s_m) \right)_j$; the constraint is violated when the LLM outputs considers the conclusion $\left( P_{n'}(s_m) \right)_j$ is false. As for true answers are highly biased towards \emph{No} answers, F1 metric between LLM outputs and the truth value of facts $\left( P_n(s_m) \right)_i$ is employed in practice.



\section{Future Research Directions}
\subsection{Extension to Complex 
Conditional and Modal Logic Reasoning}

Despite state-of-the-art approaches having greatly enhanced several logical reasoning capabilities of LLMs, there still remains a lack of exploration into the more complicated reasoning abilities such as complex conditional and modal logic reasoning. 
Specifically, sentences involving conditional reasoning can be written in the form of \emph{\textbf{If} $\ldots$, \textbf{then} $\ldots$}. 
Conditional reasoning is also related to the non-monotonic reasoning, and a recent corresponding work reveals significant limitations in current LLMs in this capability~\cite{leidinger2024llms}.
In addition, modal logic extends propositional logic by further incorporating \emph{\textbf{must}} and \emph{\textbf{might}} to account for the certainty and possibility, respectively. For example, \emph{Mary \textbf{might} ($\Diamond$) not ($\neg$) have been at the wedding (p)} implies \emph{It’s not ($\neg$) the case that Mary \textbf{must} ($\square$) have been at the wedding (p)}, which can be formally formulated as $\Diamond \neg p \vDash \neg \square p$. Despite expanding to conditional and modal logic reasoning can significantly increase the applicability of LLMs, recent work shows almost all LLMs make some basic mistakes with conditionals or modals, display logically inconsistent judgments across their inference patterns~\cite{holliday-etal-2024-conditional}. Therefore, it is still worthwhile to explore new methods to enable LLMs to reason about conditional and uncertain events.


\subsection{Higher-Order Logical Reasoning of LLMs}
Compared to first-order logic, higher-order logic (HOL) enables reasoning about properties and functions, thus facilitating more complex statements and proofs. First-order logic only quantifies over individual variables, which can only perform reasoning about properties of objects, \emph{not} properties of properties. In contrast, HOL allows quantification over functions and predicates, enabling reasoning about properties of properties and function relationships. For example, a first-order logic statement \emph{All cats are mammals} can be formulated as $\forall x (\operatorname{Cat}(x) \to \operatorname{Mammal}(x))$. A related HOL example would be \emph{There is a property that all cats have, such that any animal with that property is a mammal}, which can be formulated as $\exists P \forall x(\operatorname{Cat}(x) \to P(x)) \wedge \forall y(P(y) \to \operatorname{Mammal}(y))$. This demonstrates the ability to quantify over properties (not just individual objects), allowing for more complex expressions and reasoning on LLMs.

\subsection{Efficient Algorithms Satisfying Multiple Logical Consistencies}
While numerous methods have been proposed to enhance various types of logical consistency in LLMs, two critical challenges still remain. On the one hand, most methods are limited to enhancing a specific type of logical consistency, failing to satisfy multiple logical consistencies simultaneously. For example, the logic-guided data augmentation~\cite{asai-hajishirzi-2020-logic} only simulates reverse samples and transitive logical rules to enhance negation consistency and transitivity consistency of LLMs, respectively. Nonetheless, improving a specific type of logical consistency doesn’t necessarily enhance others. On the other hand, enumerating all possible combinations of answers to all questions to verify logical consistencies of LLMs would cost exponential space storage and unexpectedly large computational overhead. Taking the transitivity consistency checking as an example, one need to determine the answer to each of the three associated questions, which requires time complexity $\mathcal{O}(n^3)$. ZebraLogic~\cite{lin2025zebralogic} reveals LLMs' logical reasoning performance declines significantly with increasing search space and logical conflicts, named as ``curse of complexity". Efficient LLM algorithms are needed in large-scale real-world applications such as accounting~\cite{li2024applying}.
Therefore, it is essential to develop more efficient methods that can simultaneously satisfy the various logical consistencies of LLMs.

\section{Conclusion}
In summary, this survey provides a comprehensive overview of the most cutting-edge methods for enhancing LLM logical reasoning capabilities. Despite impressive advances in most NLP tasks, LLMs still face significant challenges in their logical reasoning abilities, especially in logical question answering and logical consistency. Through a thorough taxonomy, we classify state-of-the-art methods for tackling these challenges, including logical question answering via external solvers, prompting, and fine-tuning, as well as a variety of logical consistency concepts and corresponding solutions, including negation, implication, transitivity, factuality consistencies, and their composites.
In addition, we review benchmark datasets and evaluation metrics used in this research direction, which are critical for assessing LLM performance in logical reasoning tasks. 
Looking ahead, promising research directions include extending LLMs' logical reasoning abilities to modal logic for addressing questions with uncertainty and developing efficient algorithms to satisfy multiple logical consistencies simultaneously. These advancements are pivotal for enhancing the reliability and robustness of LLMs in applications requiring precise logical reasoning, and will ultimately bridge the gap between current abilities of LLMs and the demands of complex reasoning in real-world scenarios.








\newpage
\section*{Acknowledgments}
 
ZL was supported by National Key R\&D Program of China (2022ZD0160300), the Beijing Natural Science Foundation (L257007), and the NSF China (62276004). HL was supported by the NSF China (623B2002). FL was supported by the Tsinghua University Initiative Scientific Research Program. RvR was supported by the Dutch Research Council (NWO) (406.18.TW.007).

\bibliographystyle{named}
\bibliography{ijcai25}

\end{document}